\documentclass[10pt,twocolumn,letterpaper]{article}

\usepackage[pagenumbers]{iccv} 

\definecolor{iccvblue}{rgb}{0.21,0.49,0.74}
\usepackage[pagebackref,breaklinks,colorlinks,allcolors=iccvblue]{hyperref}

\usepackage[dvipsnames]{xcolor}

\definecolor{mypurple}{RGB}{135, 44, 189}
\definecolor{mygreen}{RGB}{0, 162, 67}
\definecolor{myred}{RGB}{244, 122, 31}
\definecolor{myblue}{RGB}{40, 45, 201}
\newcommand*\numcircledmod[1]{\raisebox{.5pt}{\textcircled{\raisebox{-.9pt} {#1}}}}

\usepackage{amsmath,amsfonts,bm}

\def\eqref#1{equation~\ref{#1}}

\def\1{\bm{1}}

\def\vx{{\bm{x}}}

\def\mB{{\bm{B}}}
\def\mC{{\bm{C}}}
\def\mD{{\bm{D}}}

\def\mI{{\bm{I}}}

\def\mM{{\bm{M}}}

\def\mS{{\bm{S}}}

\def\mU{{\bm{U}}}
\def\mV{{\bm{V}}}
\def\mW{{\bm{W}}}
\def\mX{{\bm{X}}}

\def\mSigma{{\bm{\Sigma}}}

\DeclareMathAlphabet{\mathsfit}{\encodingdefault}{\sfdefault}{m}{sl}
\SetMathAlphabet{\mathsfit}{bold}{\encodingdefault}{\sfdefault}{bx}{n}

\def\sC{{\mathbb{C}}}
\def\sD{{\mathbb{D}}}

\def\sM{{\mathbb{M}}}

\def\sT{{\mathbb{T}}}

\def\sW{{\mathbb{W}}}
\def\sX{{\mathbb{X}}}

\def\emA{{A}}
\def\emB{{B}}

\def\emS{{S}}

\newcommand{\R}{\mathbb{R}}

\usepackage{multirow}
\usepackage{lipsum} 
\usepackage{algorithm}
\usepackage{algpseudocode}
\def\paperID{12817} 
\def\confName{ICCV}
\def\confYear{2025}

\title{CODE-CL: \underline{Co}nceptor-Based Gradient Projection for \underline{De}ep \underline{C}ontinual \underline{L}earning}

\author{Marco~P.~E.~Apolinario \quad Sakshi~Choudhary \quad  Kaushik~Roy \\
Elmore Family School of Electrical and Computer Engineering\\
Purdue University, West Lafayete, IN 47906\\
{\tt\small mapolina@purdue.edu, choudh23@purdue.edu, kaushik@purdue.edu}
}

\begin{document}
\maketitle
\begin{abstract}
Continual learning (CL) -- the ability to progressively acquire and integrate new concepts -- is essential to intelligent systems to adapt to dynamic environments.
However, deep neural networks struggle with catastrophic forgetting (CF) when learning tasks sequentially, as training for new tasks often overwrites previously learned knowledge.
To address this, recent approaches constrain updates to orthogonal subspaces using gradient projection, effectively preserving important gradient directions for previous tasks.
While effective in reducing forgetting, these approaches inadvertently hinder forward knowledge transfer (FWT), particularly when tasks are highly correlated.
In this work, we propose \underline{Co}nceptor-based gradient projection for \underline{De}ep \underline{C}ontinual \underline{L}earning (CODE-CL), a novel method that leverages conceptor matrix representations, a form of regularized reconstruction, to adaptively handle highly correlated tasks.
CODE-CL mitigates CF by projecting gradients onto pseudo-orthogonal subspaces of previous task feature spaces while simultaneously promoting FWT. 
It achieves this by learning a linear combination of shared basis directions, allowing efficient balance between stability and plasticity and transfer of knowledge between overlapping input feature representations. 
Extensive experiments on continual learning benchmarks validate CODE-CL’s efficacy, demonstrating superior performance, reduced forgetting, and improved FWT as compared to state-of-the-art methods.\footnote{Our code is available at \href{https://github.com/mapolinario94/CODE-CL}{https://github.com/mapolinario94/CODE-CL}}

\end{abstract}    
\section{Introduction}\label{sec:introduction}
Humans possess the innate ability to continually acquire, retain and update knowledge to adapt naturally to dynamically changing environments. In contrast, while deep neural networks (DNNs) excel at leveraging massive amounts of data to generalize across various visual recognition tasks, traditional learning paradigms rely on static datasets. 
This misalignment with the ever-evolving real-world environments underscores the necessity for these models to retain past knowledge and mitigate catastrophic forgetting, as well as utilize it to enhance learning on new tasks by encouraging knowledge transfer \cite{Wang2024AApplication, Kudithipudi2022BiologicalMachines, Hadsell2020EmbracingNetworks, Ke2021FWT}. 

To address the challenges mentioned above, extensive research has focused on enabling continual learning (CL) in DNNs. 
Existing techniques fall into three categories: regularization-based, expansion-based, and memory-based methods. 
Regularization-based methods constrain updates to important model parameters for previous tasks, preserving essential features while allowing flexibility in less critical regions of the parameter space \cite{Kirkpatrick2017OvercomingNetworks, Zenke2017ContinualIntelligence, Mallya2017PackNet:Pruning, Serra2018OvercomingTask, Shi2021ContinualPreserving}. 
Expansion-based methods overcome forgetting by dynamically allocating new network resources for each task \cite{Rusu2016ProgressiveNetworks, Xu2018ReinforcedLearning, Qin2021BNS:Learning, Yoon2018LifelongNetworks, Yoon2020ScalableDecomposition}. 
Memory-based approaches, on the other hand, store representative samples for data replay or track important gradient directions from previous tasks to maintain performance on earlier data distributions \cite{Rebuffi2017ICaRL:Learning, Chaudhry2019OnLearning, Lopez-Paz2017GradientLearning, Chaudhry2019EfficientA-GEM, Zeng2019ContinualNetworks, Wang2021TrainingLearning, Farajtabar2020OrthogonalLearning, Saha2021GradientLearning}. While these techniques significantly reduce catastrophic forgetting, they inherently limit the model's ability to leverage shared information across tasks. 
In other words, they help retain past task performance, but have limited ability to utilize prior knowledge to improve learning on new tasks. 
Recent works have attempted to enhance knowledge transfer in continual learning scenarios by leveraging task similarities to integrate past knowledge into new learning\cite{Lin2022TRGP:Learning, Saha2023ContinualProjection, Lin22CUBER}. However, we demonstrate that they do not fully incorporate a systematic approach, leaving room for further improvements.

\begin{figure*}[t!]
    \begin{center}
    \includegraphics[width=\linewidth]{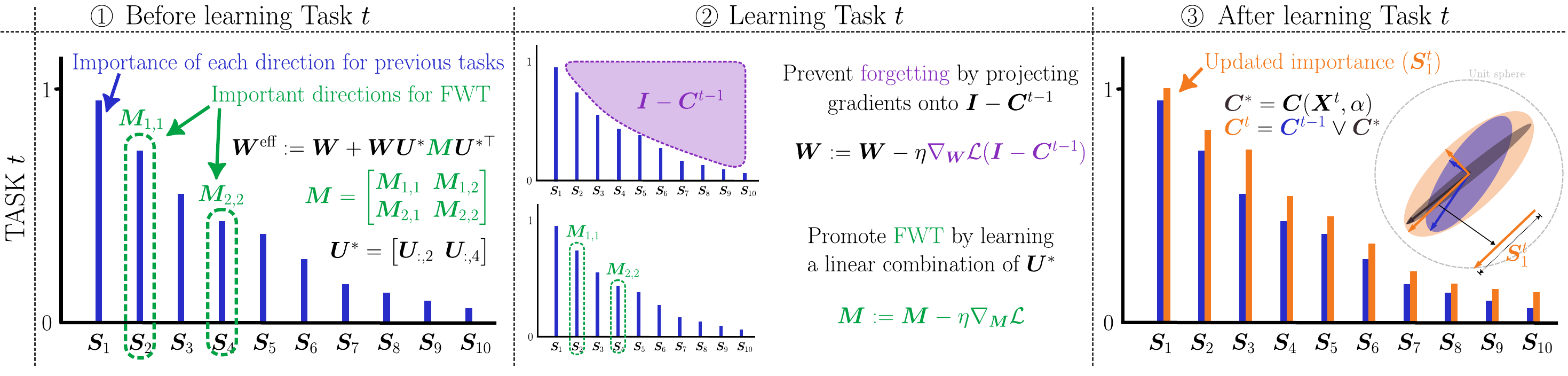}
    \end{center}
    \caption{Overview of CODE-CL. 
     \numcircledmod{1} Before learning task $t$, the {\color{myblue}importance of input activation space directions for previous tasks} is captured in the singular values {\color{myblue}$\mS_i^{t-1}$} (blue bars) of the conceptor matrix {\color{myblue}$\mC^{t-1}$} . We first identify $U^*$, the important directions for both previous tasks and the current task $t$. If such shared directions exist, we define $\mW^{\text{eff}}$ by projecting weights onto a linear combination of these common directions:  
    $\mW^{\text{eff}} = \mW + \mW\mU^*\mM\mU^{*\top}.$  
    \numcircledmod{2} During the learning phase, CODE-CL promotes {\color{mygreen}forward knowledge transfer (FWT)} by learning an optimal linear combination of the directions ({\color{mygreen}$\mM$}), while {\color{mypurple}preventing forgetting} by projecting gradients onto {\color{mypurple}$\mI - \mC^{t-1}$} (purple region). This ensures that the updates do not interfere with the previously acquired knowledge.  
    \numcircledmod{3} After learning task $t$, the {\color{myred}updated importance of each direction ($\mS^t$)} is computed by obtaining a new conceptor matrix:  
    ${\color{myred}\mC^t} = {\color{myblue}\mC^{t-1}} \lor \mC^*$, where $\mC^*$ is the conceptor matrix for task $t$. 
    Since a conceptor matrix can be interpreted as an ellipsoid in space, where its singular vectors ($\mU$) define main axes and its singular values ($\mS$) determine their lengths, operation $\lor$ corresponds to computing the minimal enclosing ellipsoid that encapsulates both conceptors. 
    In this manner, CODE-CL enables efficient continual learning by balancing knowledge retention and adaptation to new tasks.
    }
    \label{fig:method_overview}
\end{figure*}

In this paper, we propose Conceptor-Based Gradient Projection (CODE-CL), a novel continual learning algorithm that minimizes catastrophic forgetting while promoting forward knowledge transfer between tasks with highly correlated input activation subspaces. 
CODE-CL leverages conceptor matrices \cite{Jaeger2014ControllingConceptors} to enforce constrained gradient updates, preventing interference with prior tasks. More precisely, conceptor matrices provide a mathematical framework for computing the basis vectors of each layer’s input activation/feature space, which, in turn, identify the key gradient directions necessary to retain knowledge from past tasks \cite{zhang2017understanding}.
We also compute similarities between previously acquired knowledge and the new incoming task to optimize forward transfer (FWT). 
By encoding past knowledge into conceptor matrices, CODE-CL enables a structured exploration of the input activation space, allowing learning in previously restricted regions while preserving critical directions from earlier tasks. 
We give an overview of our approach in Fig.~\ref{fig:method_overview}. 
At the end of each task, CODE-CL computes a set of basis vectors, $\mU$, that span the input feature space and consequently, the gradient space for each layer using the conceptor matrix $\mC$ \cite{zhang2017understanding, Saha2021GradientLearning, Saha2023ContinualProjection}. 
For a new task, gradient updates are projected in these directions $\mathbf{U}$, scaled according to their importance for previous tasks with a regularization factor $\alpha$. 
While this constrained optimization effectively mitigates catastrophic forgetting, it does not explicitly promote forward knowledge transfer. 
To address this, we compute the intersection between the aggregated conceptor matrix and the pre-conceptor matrix of the current task, identifying common update directions $\mU^*$ and encouraging learning along these directions through $\mM$, as shown in Fig.~\ref{fig:method_overview}.

Our contributions can be summarized as follows:
\begin{itemize}
    \item We introduce CODE-CL, a novel continual learning algorithm that leverages conceptor matrices \cite{Jaeger2014ControllingConceptors} to mitigate catastrophic forgetting while effectively leveraging past learning to promote forward transfer.
    \item We evaluate the effectiveness of the proposed method through extensive experiments on standard CL vision benchmarks across various model architectures. Compared to state-of-the-art approaches, CODE-CL achieves about $1.15\%$ better final accuracy, minimal forgetting, and up to $1.18\%$ improved relative FWT.
\end{itemize}

\section{Background}\label{sec:background}
In this section, we outline the essential properties of conceptor matrices, and provide an overview of related works in continual learning.

    \subsection{Conceptor Matrices}
    Conceptor matrices constitute a mathematical framework inspired by neuroscience to encode and control the dynamics of recurrent neural networks \cite{Jaeger2014ControllingConceptors}. 
    Given a batch of feature vectors $\mX \in \R^{b \times n}$, where $b$ is the batch size and $n$ is the dimension of the feature vector space, a conceptor matrix $\mC(\mX, \alpha)$ is defined as the solution to the following minimization problem:
    \begin{equation}
        \mC(\mX, \alpha) = \arg\min_{\mC} \frac{1}{b}\| \mX - \mX\mC \|_{F}^2 + \alpha^{-2} \|\mC\|_{F}^2
    \end{equation}
    Here, $\alpha \in (0, \infty)$ is called the aperture and serves as a regularization factor. 
    This optimization problem has the following closed-form solution \cite{Jaeger2014ControllingConceptors}:  
    \begin{equation}  
    \mC(\mX, \alpha) = \frac{ \mX^\top\mX}{b} \left( \frac{ \mX^\top\mX}{b} + \alpha^{-2} \mI \right)^{-1}
    \label{eq:conceptor}
    \end{equation}
    Therefore, given the singular value decomposition (SVD) of the matrix $\mX^\top = \mU \mSigma \mV^\top$, the conceptor matrix can be expressed as $\mC = \mU \mS \mU^\top = \mU \mSigma^2 (\mSigma^2 + b\alpha^{-2}\mI)^{-1} \mU^\top$. 
    Note, the singular values of $\mC$ lie between $0$ and $1$ ($0 < \emS_{i,i} < 1$, $\forall i \in \{0, 1, \dots, n\}$), representing the importance of directions $\mU_{:,i}$. 
    In this way, $\mC$ acts as a soft projection matrix onto the linear subspace of the feature vectors of $\mX$.
    
    Conceptor matrices satisfy most laws of Boolean logic like NOT ($\neg$), OR ($\lor$), and AND ($\land$) \cite{Jaeger2014ControllingConceptors}, resulting in a simple and intuitive framework to handle the linear subspaces defined within a conceptor matrix.  
    For two conceptor matrices $\mC$ and $\mB$, we have:
    \begin{equation}
        \neg \mC = \mI - \mC
        \label{eq:not_conceptor}
    \end{equation}
    \begin{equation}
        \mC \land \mB = (\mC^{-1} + \mB^{-1} - \mI)^{-1}
        \label{eq:and_conceptor}
    \end{equation}
    \begin{equation}
        \mC \lor \mB = \neg(\neg\mC \land \neg\mB)
        \label{eq:or_conceptor}
    \end{equation}
    Here, $\neg\mC$ can be interpreted as the pseudo-orthogonal complement of the subspace characterized by $\mC$. 
    $\mC\land\mB$ signifies the conceptor matrix that describes a space that lies in the intersection between the subspaces characterized by $\mC$ and $\mB$, and $\mC\lor\mB$ describes the union between the subspaces represented by $\mC$ and $\mB$. Please refer to Section A in the Supplementary Material for additional details regarding these operations.

    We can measure the capacity, or memory usage, of a conceptor matrix based on the mean value of its singular values:
    \begin{equation}
        \Theta(\mC) = \frac{1}{n}\sum_{i=0}^n S_{i,i}
        \label{eq:conceptor_capacity}
    \end{equation}
    A capacity of $0$ would indicate that the conceptor is empty and can be represented as a null matrix, while a capacity of $1$ would indicate that the conceptor memory is full, essentially becoming an identity matrix.

    \subsection{Related Work} 
    Continual learning (CL) techniques can be broadly classified into expansion-based, regularization-based, and memory-based approaches \cite{Saha2021GradientLearning, Lin2022TRGP:Learning, Wang2024AApplication}.  
    
    Regularization-based methods mitigate forgetting by penalizing changes to important model parameters \cite{Kirkpatrick2017OvercomingNetworks, Ritter2018OnlineForgetting, Schwarz2018ProgressLearning, Zenke2017ContinualIntelligence, Serra2018OvercomingTask, Mallya2017PackNet:Pruning}. 
    While effective at preserving knowledge, these methods often rely on complex heuristics to determine parameter importance or require storing multiple model versions, leading to significant memory overhead.  
    
    Expansion-based methods address catastrophic forgetting by dynamically increasing the model's capacity as new tasks arrive \cite{Rusu2016ProgressiveNetworks, Xu2018ReinforcedLearning, Qin2021BNS:Learning, Yoon2018LifelongNetworks}. Although these approaches successfully isolate task representations to prevent interference, they result in substantial network growth, making them impractical for resource-constrained environments.  
    
    Memory-based methods mitigate forgetting by explicitly retaining information from previous tasks, either in the form of stored samples \cite{Rebuffi2017ICaRL:Learning, Chaudhry2019OnLearning} or gradient-related information \cite{Lopez-Paz2017GradientLearning, Chaudhry2019EfficientA-GEM}. 
    Within this category, orthogonal gradient projection methods \cite{He2018OvercomingBackpropagation, Zeng2019ContinualNetworks, Saha2021GradientLearning} aim to prevent interference between tasks by ensuring that the gradients are orthogonal to important directions for previous tasks.
    One such approach is Gradient Projection Memory (GPM) \cite{Saha2021GradientLearning}, which leverages the fact that gradients lie in the span of input activations \cite{zhang2017understanding}.
    Consequently, GPM utilizes singular value decomposition (SVD) on the input activations of each layer to compute and store the most important directions for each task. 
    While this prevents forgetting, it also limits forward knowledge transfer (FWT) by keeping the shared directions between old and new tasks frozen, reducing adaptability and degrading accuracy.

    Trust Region Gradient Projection (TRGP) \cite{Lin2022TRGP:Learning} addresses this by selectively allowing weight updates in a ``trusted region''. 
    Specifically, TRGP computes the projection of new task gradients onto the important directions of previous tasks and identifies the top-$ k $ tasks with the highest projections. Weight updates are then allowed along the directions associated with these $k$ tasks.
    However, this approach still lacks fine-grained adaptability, as it considers entire task subspaces rather than individual relevant directions.  
    Building on TRGP, Continual Learning with Backward Knowledge Transfer (CUBER) \cite{Lin22CUBER} introduces positive backward transfer (BWT) by 
    maintaining per-task gradient information. 
    If a new task's gradients exhibit a positive correlation with those of the previous tasks, CUBER relaxes the orthogonality constraint and introduces a regularization term to the loss function to align updates along these correlated directions.
    While this promotes positive BWT, it significantly increases memory complexity due to the need to store per-task gradients. 
    Scaled Gradient Projection (SGP) \cite{Saha2023ContinualProjection} takes a different approach by relaxing the strict orthogonality constraint of GPM. Instead of enforcing full orthogonality, SGP scales the importance of stored task directions, leading to better forward knowledge transfer (FWT) and higher average accuracy. However, SGP applies uniform scaling across all tasks, missing opportunities to adaptively exploit task similarities.
    Other notable methods include Adaptive Plasticity Improvement (API), which combines GPM's gradient constraints with dynamic model expansion when plasticity is insufficient, and Space Decoupling (SD) \cite{Zhen2023SD}, which scales gradient projections based on task correlation, allowing for a more flexible gradient update strategy compared to TRGP and GPM.  
    Taking a different perspective, Data Augmented Flatness-aware Gradient Projection (DFGP) \cite{Enneng2023DFGP} extends GPM by optimizing the loss as well as loss curvature from the perspective of both data and weights. This improves the generalization ability for new tasks and reduces catastrophic forgetting for the past tasks. However, DFGP does not explicitly leverage task similarities to facilitate knowledge transfer.
    
    In contrast to the aforementioned works, CODE-CL constrains gradients to pseudo-orthogonal directions through a regularized reconstruction framework based on conceptor matrices, as shown in (\ref{eq:conceptor}). 
    Additionally, we perform a fine-grained analysis to identify the common important directions between tasks. 
    Our proposed approach not only mitigates catastrophic forgetting but also enhances FWT by intelligently reusing prior knowledge.

\section{Methodology}\label{sec:method}

\subsection{Problem Formulation}
This work optimizes a DNN model to learn from temporally evolving data. 
We consider a supervised continual learning setting where $ T $ tasks are learned sequentially, with each task having sufficient labeled samples. 
We explore task-incremental learning scenarios in this supervised setting \cite{Wang2024AApplication}. 
Each task is identified by $ t \in \sT = \{1, 2, \dots, T\} $, and its associated dataset is represented as $ \sD^t = \{(\vx^t_i, y^t_i)_{i=1}^{n_t}\} $, where $ n_t $ is the number of samples, $ \vx^t_i $ is the input sample, and $ y^t_i $ is the corresponding label. 
Using these datasets, we train a neural network with parameters $ \sW^t = \{(\mW^{(l), t})_{l=1}^L\} $, where $ L $ represents the number of layers of the model. 
The objective is to learn parameters $ \sW^t$ such that the model performs effectively across all $T$ tasks, while mitigating catastrophic forgetting and leveraging task similarities for efficient knowledge transfer.

\begin{algorithm}[ht]
\caption{CODE-CL}\label{alg:pseudocode}
\textbf{Input:}  $\sD^t=\{(\vx^t_i, y^t_i)_{i=1}^{n_t}\}$, $\sW = \{(\mW^{(l)})_{l=1}^L\}$, aperture $\alpha$, threshold $\epsilon$, learning rate $\eta$, total training epochs $E$, number of free dimensions $K$.\\

\textbf{procedure} T\text{\scriptsize RAIN}( )\\
1. \hspace{4mm} \textbf{for} $t = 1, 2, 3, \dots, T$\\
    2. \hspace{6mm} \textbf{if} $t>1$ \textbf{then}\\
    3. \hspace{10mm} $\sX^t\gets \text{forward}(\sW^{t-1,\text{eff}}, d^t)$ for $d^t \sim \sD^t$\\
    4. \hspace{10mm} $\sC^{t, \text{pre}}\gets \text{C\scriptsize ONCEPTOR}(\sX^t, \alpha)$\\
    5. \hspace{10mm} $\sC^{t,\text{and}}\gets \sC^{t,\text{pre}}\land\sC^{t-1}$ \Comment{Equation (\ref{eq:and_conceptor})}\\
    6. \hspace{10mm} \textbf{if} $\frac{\Theta(\mC^{t,\text{and}})}{\Theta(\mC^{t-1)}}>\epsilon$ \textbf{then} \Comment{for each layer $l \in L $}\\
    7. \hspace{14mm} $\mU^{t}\gets \text{SVD}(\mC^{t, \text{and}})$\\
            8. \hspace{14mm} $\mW^{t,\text{eff}}\gets \mW(\mI+\mU^{ t}_{:,1:K}\mM^{t}\mU^{t\top}_{:,1:K})$\\
        10. \hspace{10mm} \textbf{else} \\
        11. \hspace{14mm} $\mW^{t,\text{eff}}\gets \mW$\\
        12. \hspace{10mm} \textbf{end}\\
    13. \hspace{6mm} \textbf{end}\\
    14. \hspace{6mm} $\sW^{t,\text{eff}}\gets \{(\mW^{ t,\text{eff}})_{l=1}^L\}$\\ 
    15. \hspace{6mm} \textbf{for} $e = 1, 2, 3, \dots, E$\\ 
    16. \hspace{10mm} $\nabla_{\sW}\mathcal{L}, \nabla_{\sM^t}\mathcal{L}\gets \text{SGD}(\sW^{t,\text{eff}}, d^i)$ for $d^t \sim \sD^t$\\
    17. \hspace{10mm} $\nabla_{\sW}\mathcal{L}\gets\nabla_{\sW}\mathcal{L} - \nabla_{\sW}\mathcal{L}\sC^{t-1}$ \Comment{for each layer $l$}\\ 
    18. \hspace{10mm}  $\sW \gets \sW - \eta\nabla_{\sW}\mathcal{L}$\\
    19. \hspace{10mm}  $\sM^t \gets \sM^t - \eta\nabla_{\sM^t}\mathcal{L}$\\
    20. \hspace{6mm} \textbf{end}\\
    21. \hspace{6mm} $\sX^t\gets \text{forward}(\sW, d^t)$  for $d^t \sim \sD^t$\\
    22. \hspace{6mm} $\sC^{t, \text{post}}\gets \text{C\scriptsize ONCEPTOR}(\sX^t, \alpha)$\\
    23. \hspace{6mm} \textbf{if} $t=1$ \textbf{then}\\
    24. \hspace{10mm} $\sC^t\gets \sC^{t, \text{post}}$ \\
    25. \hspace{6mm} \textbf{else}\\
    26. \hspace{10mm} $\sC^t\gets \sC^{t, \text{post}} \lor \sC^{t-1}$ \Comment{Equation (\ref{eq:or_conceptor})}\\
27. \hspace{4mm} \textbf{end}\\
\end{algorithm}

\subsection{Approach}
We demonstrate the flow of our proposed approach, CODE-CL in Algorithm~\ref{alg:pseudocode}. 
For the first task ($t=1$), learning proceeds with random weight initialization and the model is trained on dataset $ \sD^1 $ by minimizing the loss function $\mathcal{L}(\mW; \sD^1)$. 
Optimization is performed using minibatch stochastic gradient descent (SGD) without constraints. 
After training for $E$ epochs, we compute a conceptor matrix $ \mC^1 $ to encode the input subspace of each layer (lines 21-23, Algorithm~\ref{alg:pseudocode}). 
Specifically, we randomly sample $b$ inputs from $ \sD^1 $ and perform a forward pass through the model to form $ \mX^1 = [\vx^{1\top}_1, \vx^{1\top}_2, \dots, \vx^{1\top}_b] $ for each layer $l$, i.e. $\sX^1 = \{(\mX^{(l),1})_{l=1}^{L}\}$. 
Based on (\ref{eq:conceptor}), we compute the conceptor $ \mC^1 = \mC(\mX^1, \alpha)$. 

    \subsubsection{Task Overlap Analysis} \label{sec:overlap_method}
    Before training for the task $t$, we analyze the overlap between its input space and that of previous tasks, represented by $\mC^{t-1}$. 
    To do this, we forward propagate a set of inputs $\mX^t$ through the model, obtain layer-wise input activations $\sX^t$ and compute the pre-conceptor matrix $\mC^{t,\text{pre}}$ through equation (\ref{eq:conceptor}) (lines 3-4, Algorithm~\ref{alg:pseudocode}). 
    The overlap of the input space between previous tasks and the current task $t$ is represented by the intersection $ \mC^{t, \text{and}} = \mC^{t, \text{pre}} \land \mC^{t-1} $, based on (\ref{eq:and_conceptor}). 
    If many directions for the current task are encoded in $\mC^{t-1}$, tasks are highly correlated (or similar). 
    Task correlation is measured by the capacity ratio between conceptor matrices (\ref{eq:conceptor_capacity}), defining high (low) correlation when the ratio surpasses (falls below) a threshold $ \epsilon $.
    \newline \textbf{Case 1 ($\frac{\Theta(\mC^{t, \text{and} })}{\Theta(\mC^{t-1})} > \epsilon$)}: In this high correlation scenario, the directions encoded in $\mC^{t-1}$ are important for task $t$. 
    Hence, the model is allowed to learn in the top $K$ directions of $ \mC^{t, \text{and}} $ without negatively impacting prior tasks. 
    To achieve this, the weights are projected onto the subspace defined by these directions as follows:
        \begin{equation}
            \mW^{t, \text{eff}}=\mW + \mW \mU^{t, \text{and}}_{:, 1:K}\mM^{t}\mU^{t, \text{and}\top}_{:, 1:K},
        \end{equation}
    where $\mU^{t, \text{and}}_{:, 1:K}$ are the top-$K$ singular vectors of $\mC^{t, \text{and}}$, and $\mM \in \R^{K \times K}$ is a task-specific learnable matrix which defines the extent of learning in these directions. 
    This formulation explicitly allows us to utilize past knowledge to improve the performance of the current task, thereby improving forward knowledge transfer.
    \newline \textbf{Case 2 ($\frac{\Theta(\mC^{t, \text{and} })}{\Theta(\mC^{t-1})} \leq \epsilon$):} In this case, the task overlap is minimal, leaving little possibility of forward transfer. Thus, the effective weights remain $ \mW^{t, \text{eff}} = \mW $.
    
    \subsubsection{Constrained Gradient Updates}
    While learning task $t$, the model is trained on $\sD^t$ to minimize the following loss function:
        \begin{equation}
            \begin{split}
                \mW^t, \mM^t &:= \arg\min_{\mW, \mM} \mathcal{L}(\mW^{t, \text{eff}}; \sD^t) \\ 
                & \text{s.t.} \nabla_{\mW}\mathcal{L} =\nabla_{\mW}\mathcal{L}(\mI-\mC^{t-1}) 
            \end{split},
        \end{equation}
    where the gradients are constrained to lie in the pseudo-orthogonal subspace of the conceptor matrix defined by $\neg \mC^{t-1}$ (\ref{eq:not_conceptor}), where $\mC^{t-1}$ contains important directions for previous tasks, scaled by aperture $\alpha$, as shown in (\ref{eq:conceptor}).    
    \subsubsection{Post Training Conceptor Update}
    After training for task $t$,  we merge the current and past task knowledge into a new conceptor matrix $\mC^t$ for each layer (line 26, Algorithm~\ref{alg:pseudocode}). 
    This is achieved by first computing the post-training conceptor matrix $\mC^{t, \text{post}}$, as shown in lines 21-22 in Algorithm~\ref{alg:pseudocode}. 
    We then merge $ \mC^{t, \text{post}} $ and $ \mC^{t-1} $ into a new conceptor matrix, consolidating the important directions for all learned tasks based on (\ref{eq:or_conceptor}).

\section{Experiments}\label{sec:experimental}
In this section, we first provide details regarding our experimental setup, and then show the efficacy of CODE-CL through extensive experiments across various continual learning benchmarks.

    \begin{table*}[!htbp]
        \caption{Performance comparison on continual image classification datasets using multi-head networks. Accuracy and BWT (mean $\pm$ std) are reported over five trials. Best results are in bold and second best are underlined. \textsuperscript{\dag} denotes the results taken from \cite{Saha2021GradientLearning} and \textsuperscript{\ddag} denote the results from the respective original papers. All other results are reproduced based on their official open source implementations.}
        \label{table:continual_image}
        \centering
        \resizebox{0.9\textwidth}{!}{
        \begin{tabular}{l|llllll}
        \toprule
        \multicolumn{1}{c|}{\multirow{2}{*}{Method}}  &
          \multicolumn{2}{c}{Split CIFAR100} &
          \multicolumn{2}{c}{Split MiniImageNet} &
          \multicolumn{2}{c}{5-Datasets} 
          \\ \cmidrule{2-7} 
        \multicolumn{1}{c|}{} &
          \multicolumn{1}{c}{ACC (\%)} &
          \multicolumn{1}{c}{BWT (\%)} &
          \multicolumn{1}{c}{ACC (\%)} &
          \multicolumn{1}{c}{BWT (\%)} &
          \multicolumn{1}{c}{ACC (\%)} &
          \multicolumn{1}{c}{BWT (\%)} 
          \\
        \midrule
        Multitask\textsuperscript{\dag} & $79.58 \pm 0.54$ & $-$ & $69.46 \pm 0.62$ & $-$ & $91.54 \pm 0.28$ & $-$ 
        \\
        \midrule
        OWM \cite{Zeng2019ContinualNetworks}\textsuperscript{\dag} & $50.94 \pm 0.60$ & -$30 \pm 1$ & $-$ & $-$ & $-$ & $-$ 
        \\
        EWC \cite{Kirkpatrick2017OvercomingNetworks}\textsuperscript{\dag} & $68.80 \pm 0.88$ & -$2 \pm 1$ & $52.01 \pm 2.53$ & -$12 \pm 3$ & $88.64 \pm 0.26$ & -$4 \pm 1$ 
        \\
        HAT \cite{Serra2018OvercomingTask}\textsuperscript{\dag} & $72.06 \pm 0.50$ & $0 \pm 0$ & $59.78 \pm 0.57$ & -$3 \pm 0$ & $91.32 \pm 0.18$ & -$1\pm0$ 
        \\
        A-GEM \cite{Chaudhry2019EfficientA-GEM}\textsuperscript{\dag} & $63.98 \pm 1.22$ & -$15 \pm 2$ & $57.24 \pm 0.72$ & -$12 \pm 1$ & $84.04 \pm 0.33$ & -$12 \pm 1$ 
        \\
        ER\_Res \cite{Chaudhry2019OnLearning}\textsuperscript{\dag} & $71.73 \pm 0.63$ & -$6 \pm 1$ & $58.94 \pm 0.85$ & -$7 \pm 1$ & $80.31 \pm 0.22$ & -$4 \pm 0$ 
        \\
        API \cite{Yan2023API}\textsuperscript{\ddag} & $-$ & $-$ & $65.9\pm0.6$ & -$0.3\pm0.2$ & $91.1\pm0.3$ & -$0.5\pm0.1$ \\
        DFGP \cite{Enneng2023DFGP}\textsuperscript{\ddag} & $74.59\pm0.33$ & -$0.9$ & \underline{$69.92\pm0.9$} & -$1$ & $92.09\pm0.18$ & -$1$ \\
        TRGP+SD \cite{Zhen2023SD}\textsuperscript{\ddag} & $75.50\pm0.35$ & -$2.88\pm0.89$ & $65.8\pm0.16$ & -$0.49\pm0.08$ & $-$ & $-$ \\
        GPM \cite{Saha2021GradientLearning} &
          $72.06\pm0.29$ &
          -$0.2\pm0.19$ &
          $66.26\pm1.18$ &
          -$0.9\pm1.34$ &
          $90.70\pm0.45$ &
          -$1.0\pm0.16$ 
          \\
        TRGP \cite{Lin2022TRGP:Learning} &
          $75.24\pm0.29$ &
          -$0.1\pm0.18$ &
          $65.08\pm0.94$ &
          -$0.5\pm0.74$ &
          \underline{$92.81\pm0.54$} &
          -$0.1\pm0.03$ 
          \\
        CUBER \cite{Lin22CUBER} &
          $75.30\pm0.43$ &
          $0.1\pm0.11$ &
          $64.25\pm0.75$ &
          -$0.7\pm0.48$ &
          $92.77\pm0.60$ &
          -$0.03\pm0.02$ 
          \\
        SGP \cite{Saha2023ContinualProjection} &
          \underline{$75.69\pm0.38$} &
          -$1.4\pm0.17$ &
          $68.50\pm2.09$ &
          -$2.0\pm2.10$ &
          $90.42\pm0.66$ &
          -$1.61\pm0.31$ 
          \\
        \midrule
        CODE-CL (Ours) &
          $\mathbf{77.21\pm0.32}$ &
          -$1.1\pm0.28$ &
          $\mathbf{71.16\pm0.32}$ &
          -$1.1\pm0.3$ &
          $\mathbf{93.51\pm0.13}$ &
          -$0.11\pm0.01$ 
          \\
        \bottomrule
        \end{tabular}
        }
    \end{table*}

    \subsection{Experimental Setup}\label{sec:experimental_setup}
    In this subsection, we outline the benchmarks, network architectures, training hyperparameters, and performance metrics used to evaluate and compare our method with state-of-the-art CL techniques.
    
    \subsubsection{Benchmarks and Models}
    We evaluate our method on widely used continual learning (CL) benchmarks, including Split CIFAR100 \cite{Krizhevsky2009LearningImages}, Split miniImageNet \cite{Vinyals2016MatchingLearning}, and 5-Datasets \cite{Ebrahimi2020AdversarialLearning}. 
    For Split CIFAR100, the original CIFAR100 dataset is divided into $T$ groups, each containing an equal number of classes ($100/T$). 
    In our experiments, we split the dataset into 10 groups, with each group representing a separate task, and train a 5-layer AlexNet model in a multi-head setting, where each head is associated with one unique task \cite{Saha2021GradientLearning, Lin2022TRGP:Learning, Saha2023ContinualProjection}. 
    Similarly, the Split miniImageNet benchmark consists of a subset of 100 classes from the ImageNet dataset, divided into 20 groups. 
    The 5-Datasets benchmark involves training a model sequentially on five different datasets: CIFAR10, MNIST, SVHN, notMNIST, and Fashion MNIST. For both Split miniImageNet and 5-Datasets, we use a reduced ResNet18 model in a multi-head setting \cite{Saha2021GradientLearning, Lin2022TRGP:Learning, Saha2023ContinualProjection}. To ensure a fair comparison with prior works, we refrain from using data augmentation in our experiments. 
    The dataloaders for Split CIFAR100 and 5-Datasets are obtained from GPM\cite{Saha2021GradientLearning}, while the one for Split miniImageNet was provided by the Avalanche library \cite{Carta2023Avalanche:Learning}.
    
    \subsubsection{Training Details}
    For all our experiments, we use stochastic gradient descent (SGD) with a learning rate scheduler and early stopping criteria \cite{Saha2021GradientLearning}.
    Each task in Split CIFAR100 is trained for a maximum of 200 epochs with a batch size of 64 and aperture $\alpha=6$. Similarly, each task in Split miniImageNet and 5-Datasets is trained for a maximum of 100 epochs with a batch size of 64, and with $\alpha=16$ and $\alpha=8$, respectively.
    For all our experiments as shown in Table~\ref {table:continual_image}, we use $K=80$.
    For more details on our implementation, please refer to Section C in the Supplementary Material.

    \subsubsection{Performance Metrics}
    Similar to previous works \cite{Lopez-Paz2017GradientLearning, Saha2021GradientLearning, Lin2022TRGP:Learning, Saha2023ContinualProjection}, we use three metrics to evaluate the performance of our method: the average final accuracy over all tasks, Accuracy (ACC), Backward Transfer (BWT), which measures the forgetting of old tasks when learning new tasks, and relative Forward Transfer (FWT), which measures the beneficial effects of learning the previous tasks for learning a new one. 
    ACC and BWT are defined as:
    \begin{equation}
        \text{ACC} = \sum_{i=1}^{T} \frac{\emA_{T,i}}{T} \text{ ; }
        \text{BWT} = \sum_{i=1}^{T-1} \frac{\emA_{T,i}  - \emA_{i,i}}{T-1},
    \end{equation}
    where $T$ is the number of tasks, $\emA_{j,i}$ is the accuracy of the model on $i$-th task after learning the $j$-th task sequentially ($i\leq j$). 
    Similarly, FWT is defined as:
     \begin{equation}
        \text{FWT} = \frac{1}{T}\sum_{i=1}^{T} \emA_{i,i}  - \emB_{i,i},
    \end{equation}
    where $\emB_{i,i}$ is the accuracy of a baseline method used for training the same model on the $i$-th task.
    In our experiments, we used GPM \cite{Saha2021GradientLearning} as the baseline to compare other methods.

    \subsection{Results}
    Here, we present the performance of CODE-CL in comparison with prior approaches, along with a detailed analysis of its memory complexity. 
    Additionally, we conduct ablation studies to assess the impact of varying the number of free dimensions, $K$, on the method's performance.

    \begin{table}[!htbp]
        \caption{Comparison of relative FWT with respect to GPM \cite{Saha2021GradientLearning}. Values (mean $\pm$ std) are reported over five trials. Best results are in bold and second best are underlined.}
        \label{table:fwt_comparison}
        \centering
        \resizebox{\columnwidth}{!}{
        \begin{tabular}{l|ccc}
        \toprule
        \multirow{2}{*}{Method}  &
          \multicolumn{1}{c}{S-CIFAR100} &
          \multicolumn{1}{c}{S-MiniImageNet} &
          \multicolumn{1}{c}{5-Datasets} 
          \\ \cmidrule{2-4} 
        \multicolumn{1}{c|}{} &
          \multicolumn{1}{c}{FWT (\%)} &
          \multicolumn{1}{c}{FWT (\%)} &
          \multicolumn{1}{c}{FWT (\%)} 
          \\
        \midrule
        TRGP \cite{Lin2022TRGP:Learning} &
          $2.86\pm0.26$ &
          -$1.56\pm0.67$ &
          $\underline{1.16\pm0.52}$ 
          \\
        CUBER \cite{Lin22CUBER} &
          $2.86\pm0.49$ &
          -$2.22\pm0.70$ &
          $1.10\pm0.60$ 
          \\
        SGP \cite{Saha2023ContinualProjection} &
          $\underline{4.74\pm0.37}$ &
          $\underline{3.37\pm0.88}$ &
          $0.33\pm0.37$ 
          \\
        \midrule
        CODE-CL &
          $\bm{5.92\pm0.34}$ &
          $\bm{4.17\pm0.41}$ &
          $\bm{1.82\pm0.12}$ 
          
          \\
        \bottomrule
        \end{tabular}
        }
    \end{table}

    \begin{table*}[!ht]
        \caption{Memory complexity comparison among methods. The analysis is done for a single fully-connected layer with $N$ inputs, $M$ outputs, after being trained on $T$ tasks. Also, $B$ is the average number of important direction per task used in \cite{Lin2022TRGP:Learning} and $K$ is the number of free dimensions parameter used in CODE-CL.}
        \label{table:memory}
        \centering
        \resizebox{0.9\textwidth}{!}{
        \begin{tabular}{l|ccccc}
        \toprule
        Methods           & GPM & TRGP & CUBER & SGP & CODE-CL (Ours) \\ 
        \midrule
        Memory Complexity & $O(N^2)$ &  $O(N^2 + TNB + TB^2)$ & $O(N^2 + TN^2 + TNB + TB^2)$ & $O(N^2)$ & $O(N^2 + TNK + TK^2)$               \\ 
        \bottomrule
        \end{tabular}
        }
    \end{table*}

    \subsubsection{Performance Comparison} 
    As shown in Table~\ref{table:continual_image}, our method achieves high accuracy with minimal forgetting across all benchmarks. 
    Specifically, CODE-CL consistently delivers competitive results, outperforming previous methods on all three datasets.  
    
    In terms of accuracy, on Split CIFAR100, CODE-CL achieves $77.21\%$, coming close to the upper bound set by Multitask Learning ($79.58\%$), which serves as an ideal but unrealistic comparison point. 
    Notably, CODE-CL outperforms other state-of-the-art continual learning methods, achieving higher accuracy than all previous approaches, including API, DFGP, TRGP+SD, GPM, TRGP, CUBER, and SGP. 
    Similarly, on Split MiniImageNet and 5-Datasets, CODE-CL once again performs exceptionally well, surpassing all other previous methods. 
    In both cases, it even exceeds the accuracy of the Multitask Learning baseline, illustrating the beneficial effect of forward knowledge transfer when learning tasks sequentially. 
    This further underscores CODE-CL’s robustness, particularly on more challenging datasets, where competing methods tend to suffer significant performance drops.  

    \begin{figure}[t!]
    \begin{center}
    \includegraphics[width=0.9\linewidth]{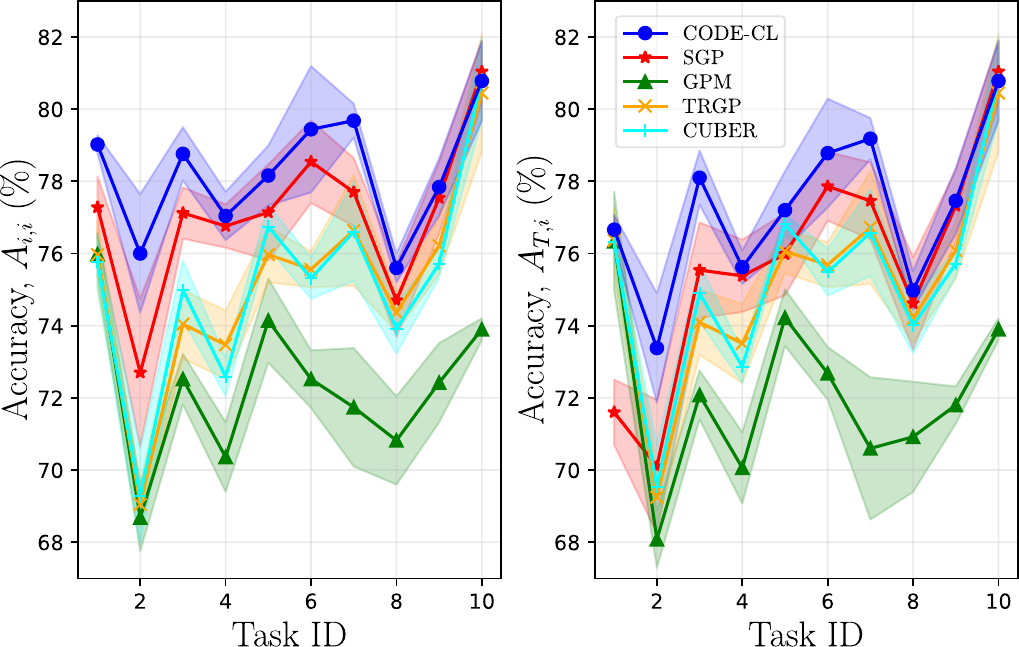}
    \end{center}
    \caption{Test accuracy of each task on the Split CIFAR100 benchmark: (left) immediately after learning the task, $\emA_{i,i}$; (right) after learning all tasks, $\emA_{T,i}$. Here, it can be seen that our method outperforms previous methods for all tasks.}
    \label{fig:cifar100_evolution}
    \end{figure}  
    
    Fig.~\ref{fig:cifar100_evolution} presents the model’s accuracy for each task immediately after learning it ($\emA_{i,i}$) and after sequentially learning all tasks ($\emA_{T,i}$) in the Split CIFAR100 dataset. 
    The difference between these two measures quantifies the extent of forgetting. As shown, CODE-CL achieves superior $\emA_{i,i}$ and $\emA_{T,i}$ compared to other methods across all tasks. 
    This advantage arises mainly because, unlike methods such as GPM, TRGP, or CUBER, CODE-CL incorporates pseudo-orthogonal gradient projections. 
    Additionally, in contrast to SGP, our method enables the selective release of important shared directions, further enhancing forward transfer.  
    
    To quantify this, we measure the relative FWT of key representative methods (TRGP, CUBER, and SGP) and compare them against CODE-CL using GPM as a reference. 
    The results, presented in Table~\ref{table:fwt_comparison}, demonstrate that CODE-CL consistently achieves better FWT. 
    This can be attributed to its relaxation of gradient projections into pseudo-orthogonal spaces, unlike TRGP or CUBER, and its fine-grained selection of the most important shared directions among tasks, unlike the other methods.  
    
    In terms of BWT, our results further illustrate CODE-CL's effectiveness in mitigating catastrophic forgetting. 
    On Split CIFAR100, CODE-CL records a BWT of -$1.1\%$, indicating minimal performance loss on previously learned tasks, comparable to prior works. 
    Similarly, on Split MiniImageNet, CODE-CL achieves a BWT of -$1.1\%$, aligning with state-of-the-art methods and demonstrating its ability to retain learned knowledge with minimal degradation.
    Finally, on the 5-Datasets benchmark, CODE-CL reports a BWT of -$0.11\%$, performing similarly to TRGP.  
    
    In summary, the high accuracy, low forgetting, and improved FWT of CODE-CL highlight its ability to effectively balance the trade-off between plasticity and stability, maintaining strong performance across a range of continual learning tasks while minimizing forgetting.

    \subsubsection{Memory Complexity}
    We analyze the memory complexity of our proposed approach and compare it with state-of-the-art techniques GPM, SGP, TRGP and CUBER.
For simplicity, we analyze a single fully connected layer with $N$ inputs and $M$ outputs after training on $T$ tasks. 
    CODE-CL’s memory complexity is primarily influenced by conceptor matrices of size $N^2$, which encode input vector space information. 
    Additionally, as discussed in Section~\ref{sec:method}, CODE-CL allocates a fixed number of free dimensions per task ($K$) to learn an optimal linear combination of the $K$ most important directions within the subspace formed by the intersection of past and new task conceptors. 
    This introduces an additional memory requirement of $TNK + TK^2$, where $TNK$ accounts for the storage of $K$ key directions per task of dimension $N$, and $TK^2$ accounts for learnable square matrices $\mM^{t}$. 
    Consequently, the total memory complexity of CODE-CL is $O(N^2 + TNK + TK^2)$.
    
    For GPM and SGP, memory usage is determined solely by the input dimension $N$, leading to a $O(N^2)$ complexity. 
    TRGP shares this base complexity but also stores important directions per task and trusted region projection subspaces, incurring an additional cost of $O(TNB + TB^2)$, where $B$ is the number of important directions per task. 
    Similarly, CUBER requires $O(N^2 + TN^2 + TNB + TB^2)$, with the additional $TN^2$ term arising from additional gradient storage needs.
    
\begin{figure}[t!]
    \begin{center}
    \includegraphics[width=0.9\linewidth]{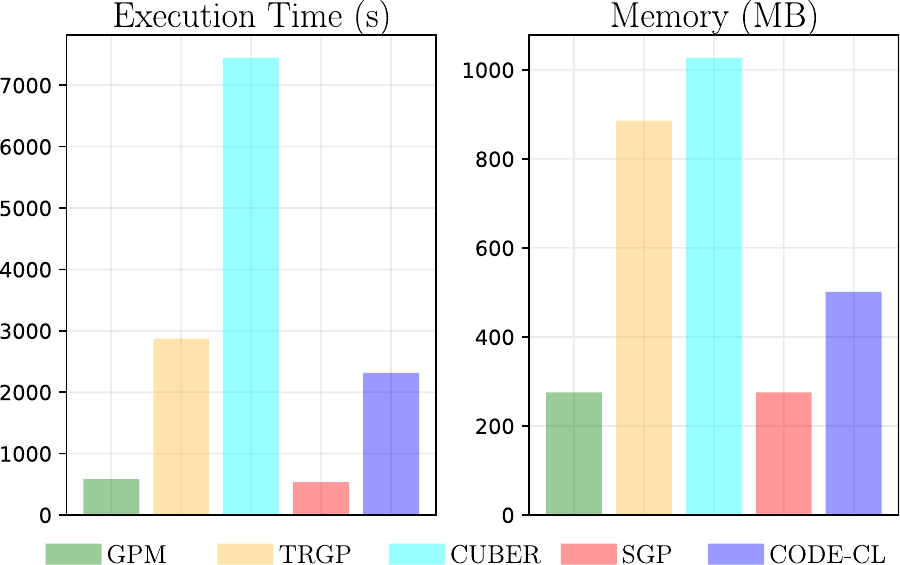}
    \end{center}
    \caption{Execution time (left) and memory (right) comparison on the Split CIFAR100 benchmark. Lower means better. Here, CODE-CL represent a more efficient method than techniques such as TRGP or CUBER.  }
    \label{fig:time_mem_stats}
    \end{figure}

    Table~\ref{table:memory} summarizes the memory complexity of each method. 
    In particular, as model size (i.e. $N$) grows, CODE-CL maintains a fixed and significantly smaller number of free dimensions ($K \ll N$), making its memory requirements comparable to GPM, SGP, and TRGP, while being significantly lower than CUBER.
    GPU memory usage measurements on Split CIFAR100 (Fig.~\ref{fig:time_mem_stats}) confirm CODE-CL's efficiency, requiring half the memory of TRGP and CUBER. 
    Additionally, CODE-CL achieves a slightly shorter execution time than TRGP and is approximately $3\times$ faster than CUBER. 
    While it introduces some overhead compared to GPM and SGP, this trade-off is justified by its superior performance.

    \subsubsection{Ablation Study}
    In this section, we examine the impact of the number of free dimensions ($K$) and the aperture parameter ($\alpha$) on performance. 
    Note that we modify only one parameter at a time, keeping all other training hyperparameters fixed.
    
    \begin{figure}
    \centering
    \begin{subfigure}{0.23\textwidth}
    \includegraphics[height=5.0cm]{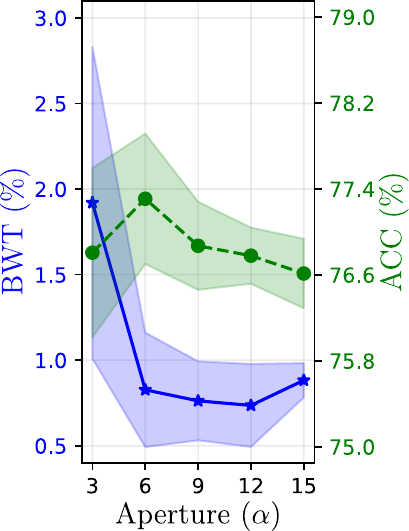}
    \caption{Split CIFAR100}
    \end{subfigure}
    \hfill
    \begin{subfigure}{0.23\textwidth}
    \includegraphics[height=5.0cm]{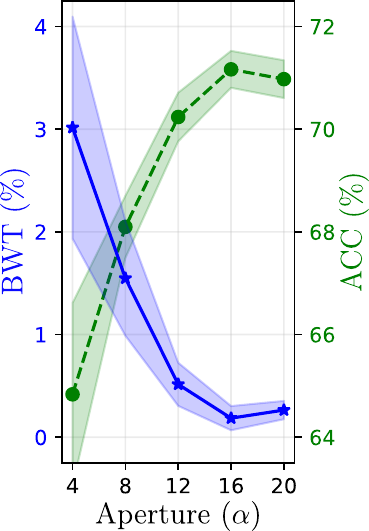}
    \caption{Split MiniImageNet}
    \end{subfigure}
    \caption{Effect of the aperture ($\alpha$) parameter on ACC and BWT for the Split CIFAR-100 and Split miniImageNet benchmarks. In both cases, results show that the greater the $\alpha$ ($\uparrow$) parameter, the lower the BWT ($\downarrow$), meaning the model forgets less.}
    \label{fig:alpha_ablation}
    \end{figure}

    \begin{figure}
    \centering
    \begin{subfigure}{0.23\textwidth}
    \includegraphics[height=5.0cm]{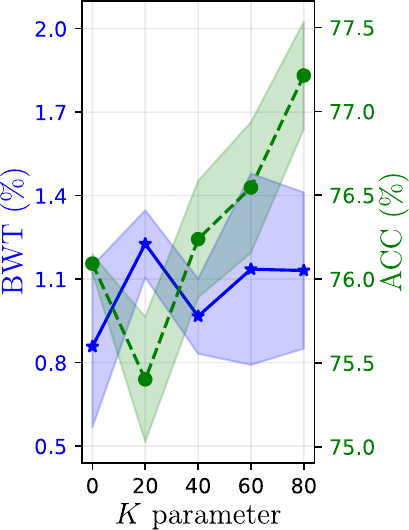}
    \caption{Split CIFAR100}
    \end{subfigure}
    \hfill
    \begin{subfigure}{0.23\textwidth}
    \includegraphics[height=5.0cm]{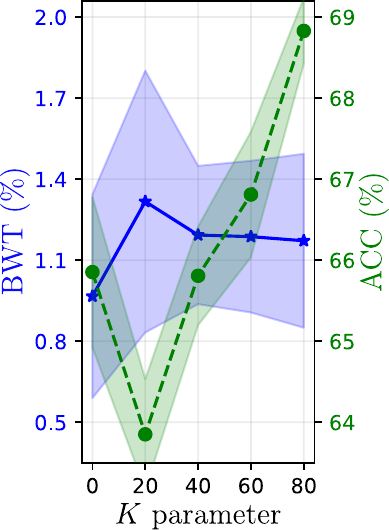}
    \caption{Split MiniImageNet}
    \end{subfigure}
    \caption{Effect of the number of free dimensions ($K$) on the final accuracy and BWT for the Split CIFAR-100 and Split miniImageNet benchmarks. In both cases, results show that for $K > 20$, the greater the $K$ ($\uparrow$), the greater the ACC ($\uparrow$), while BWT does not change significantly with $K$.}
    \label{fig:k_ablation}
    \end{figure}

    \paragraph{Effects $\alpha$ in performance:}
    As shown in Fig.~\ref{fig:alpha_ablation}, $\alpha$ directly influences the model's forgetting rate. 
    Specifically, higher values of $\alpha$ bring BWT closer to zero, meaning the model forgets less. 
    This behavior aligns with the definition of conceptor matrices, as $\alpha$ scales the singular values of the data. When $\alpha\rightarrow\infty$, the conceptor matrices approximate the identity matrix, preventing forgetting entirely. 
    However, this also means that the model loses plasticity and is unable to integrate new information. 
    This trade-off is evident in Fig.~\ref{fig:alpha_ablation}: for Split CIFAR100, peak performance is achieved at $\alpha=6$, whereas for Split MiniImageNet, the optimal value is $\alpha=16$.
    
    \paragraph{Effects $K$ in performance:}
    The results for both benchmarks, presented in Fig.~\ref{fig:k_ablation}, indicate that increasing $K$ generally improves accuracy while maintaining low BWT. 
    When $K > 20$, higher values of $K$ lead to greater ACC, suggesting that increasing $K$ enhances overall performance by facilitating forward knowledge transfer from previous tasks to new ones. 
    However, its impact on BWT reduction remains minimal. 
    While increasing $K$ may seem advantageous, it comes with additional memory overhead, making it crucial to balance performance gains with memory efficiency.

    \subsubsection{Comparison on Tasks with Overlapping Classes}
    Most of the benchmarks used in this study consist of tasks with non-overlapping classes, although they share similarities in the feature space, as reflected in neuronal activity representations. 
    While these benchmarks effectively demonstrate CODE-CL’s ability to identify the most relevant directions in overlapping feature spaces, evaluating our method on a benchmark with overlapping classes can further highlight its advantages.
    To this end, we adopted the OL-CIFAR100 benchmark \cite{Lin22CUBER}, where the first 50 classes of CIFAR100 are split into seven tasks. 
    Specifically, Tasks 0–6 contain the following class distributions: 0–9, 5–14, 10–19, 20–29, 25–34, 30–39, and 40–49, respectively. 
    
    The results of this evaluation are summarized in Table~\ref{table:ol_cifar100}.
    Here, CODE-CL outperforms previous methods in terms of ACC, demonstrating the benefits of our approach in scenarios with class overlap. 
    Additionally, we compute the relative FWT with respect to GPM. The superior FWT of CODE-CL underscores the effectiveness of our fine-grained selection of important directions within overlapping input feature subspaces. 
    This, combined with pseudo-orthogonal gradient updates, leads to more efficient forward transfer learning compared to methods like TRGP or CUBER, which rely on full task directions, or SGP, which only considers pseudo-orthogonal gradient updates.
    
    \begin{table}[!htbp]
        \caption{Comparison of methods performance on OL-CIFAR100. Values (mean $\pm$ std) are reported over five trials. Best results are in bold and second best are underlined.}
        \label{table:ol_cifar100}
        \centering
        \resizebox{\columnwidth}{!}{
        \begin{tabular}{l|ccc}
        \toprule
        \multirow{1}{*}{Method}  &
          \multicolumn{1}{c}{ACC (\%)} &
          \multicolumn{1}{c}{BWT (\%)} &
          \multicolumn{1}{c}{FWT (\%)} 
          \\
        \midrule
        GPM \cite{Saha2021GradientLearning} &
          $71.62\pm0.45$ &
          -$0.34\pm0.15$ &
          $0$ 
          \\
        TRGP \cite{Lin2022TRGP:Learning} &
          $74.77\pm0.43$ &
          -$0.06\pm0.10$ &
          $2.73\pm0.34$ 
          \\
        CUBER \cite{Lin22CUBER} &
          \underline{$75.01\pm0.23$} &
          -$0.01\pm0.26$ &
          $3.02\pm0.20$ 
          \\
        SGP \cite{Saha2023ContinualProjection} &
          $75.00\pm0.68$ &
          -$1.75\pm0.59$ &
          \underline{$4.79\pm0.42$} 
          \\
        \midrule
        CODE-CL &
          $\mathbf{76.89\pm0.42}$ &
          -$1.01\pm0.18$ &
          $\mathbf{6.02\pm0.36}$ 
          
          \\
        \bottomrule
        \end{tabular}
        }
    \end{table}
    
\section{Conclusion}\label{sec:conclusions}
We introduce CODE-CL, a novel continual learning algorithm that leverages conceptor matrices to mitigate catastrophic forgetting while enhancing forward transfer. 
CODE-CL achieves this by projecting gradients onto pseudo-orthogonal subspaces of previous task feature spaces and learning a linear combination of shared basis directions. 
This approach effectively balances stability and plasticity, allowing efficient knowledge transfer across overlapping feature representations. 
Extensive experiments on standard continual learning benchmarks demonstrate CODE-CL’s effectiveness, achieving superior accuracy, minimal forgetting, and improved forward transfer compared to state-of-the-art methods.

\section*{Acknowledgments}
This work was supported in part by the Center for Co-design of Cognitive Systems (CoCoSys), one of the seven centers in JUMP 2.0, a Semiconductor Research Corporation (SRC) program, and in part by the Department of Energy (DoE).

{
    \small
    \bibliographystyle{ieeenat_fullname}
    \bibliography{references}
}
\clearpage
\setcounter{page}{1}
\maketitlesupplementary
\appendix
\renewcommand{\algorithmicrequire}{\textbf{Input:}}
\renewcommand{\algorithmicensure}{\textbf{Output:}}

\section{Conceptor Implementation Details}\label{appendix:conceptor}

We implement the conceptor operations following the equations presented in Section~\ref{sec:background}, with one exception: the AND operation (\ref{eq:and_conceptor}). 

The operation defined in (\ref{eq:and_conceptor}) is only valid when the conceptor matrices are invertible. 
However, in practice, since we use a limited number of samples to compute the conceptors, the resulting matrices are often not full rank. 
To address this, we adopt a more general version of the AND operation, as proposed in \cite{Jaeger2014ControllingConceptors}:
\begin{equation}
    \mC \land \mB = \mD (\mD^\top (\mC^\dagger + \mB^\dagger - \mI)\mD)^{-1}\mD^\top,
    \label{eq:and_general_conceptor}
\end{equation}
Here, $\mC^{\dagger}$ and $\mB^{\dagger}$ denote the pseudo-inverses of $\mC$ and $\mB$, respectively. 
The matrix $\mD$ consists of columns that form an arbitrary orthonormal basis for the intersection of the column spaces of $\mC$ and $\mB$. 

The procedure for computing $\mD$ is outlined in Algorithm~\ref{alg:pseudocode_dcomputation}.

\begin{algorithm}
\caption{Computation of matrix $\mD$ in (\ref{eq:and_general_conceptor})}\label{alg:pseudocode_dcomputation}
\begin{algorithmic}
\Require $\mC$, $\mB$, $\beta$ (threshold), $N$ (dimension of $\mC$ and $\mB$)
\Ensure $\mD$
\State $\mU_\mC, \mS_\mC \gets \text{SVD}(\mC)$    \Comment{Singular value decomposition}
\State $\mU_\mB, \mS_\mB \gets \text{SVD}(\mB)$ 
\State $k_{\mC}\gets \text{num\_elements}(\mS_{\mC}>\beta)$ \Comment{\# of elements $>\beta$}
\State $k_{\mB}\gets \text{num\_elements}(\mS_{\mB}>\beta)$
\State $\mU_\mC' \gets \mU_\mC[:, k_\mC:]$  \Comment{Last $N-k_\mC$ columns}
\State $\mU_\mB' \gets \mU_\mB[:, k_\mB:]$ 
\State $\mU, \mS\gets \text{SVD}(\mU_\mC'\mU_\mC'^\top+\mU_\mB'\mU_\mB'^\top)$
\State $k\gets \text{num\_elements}(\mS>\beta)$
\State $\mD \gets \mU[:, k:]$

\end{algorithmic}
\end{algorithm}
\section{Additional Ablation Studies}
In this section, we present additional ablation studies to evaluate the impact of the number of free dimensions ($K$) and aperture ($\alpha$) on the 5-Datasets benchmark, as well as the effect of the threshold parameter ($\epsilon$) across all three benchmarks.

Tables~\ref{table:ablation_alpha_5} and \ref{table:ablation_k_5} summarize the results on the 5-Datasets benchmark. 
We observe that increasing $\alpha$ leads to a reduction in BWT, consistent with the findings in Section~\ref{sec:experimental}. 
Similarly, increasing $K$ improves final accuracy, further validating trends observed in the other datasets.

Regarding the threshold parameter ($\epsilon$), results suggest that lower values of $\epsilon$ enhance performance by allowing more directions in the intersection of input spaces across tasks to be freed. 
However, this also increases memory requirements. 
Therefore, selecting an appropriate $\epsilon$ involves a trade-off between performance and computational resources.

\section{Experimental Setup}\label{appendix:setup}

This section provides details on the architecture of all models used in this work, the dataset statistics, the hyperparameters for each experiment, and the compute resources employed.

\begin{table}[!t]
    \caption{Ablation studies on the aperture ($\alpha$) hyperparameter on the 5-Datasets benchmark. Results are reported as mean $\pm$ standard deviation over five trials. Other hyperparameters are constant as reported in Section~\ref{sec:experimental}.}
    \label{table:ablation_alpha_5}
    \centering
    \resizebox{0.6\columnwidth}{!}{%
    \begin{tabular}{c|ll}
    \toprule
     $\alpha$  & \multicolumn{1}{c}{ACC ($\%$)} & \multicolumn{1}{c}{BWT ($\%$)} \\
     \midrule
     $4$ & $93.32\pm0.13$ & $-0.25\pm0.02$\\
     $8$ & $\mathbf{93.51\pm0.13}$ & $-0.11\pm0.01$\\
     $16$ & $93.46\pm0.16$ & $-0.04\pm0.00$\\
     \bottomrule
    \end{tabular}%
    }
    \end{table}

    \begin{table}[!t]
    \caption{Ablation studies on the number of free dimensions ($K$) parameter on the 5-Datasets benchmark. Results are reported as mean $\pm$ standard deviation over five trials. Other hyperparameters are constant as reported in Section~\ref{sec:experimental}.}
    \label{table:ablation_k_5}
    \centering
    \resizebox{0.6\columnwidth}{!}{%
    \begin{tabular}{c|ll}
    \toprule
     $K$  & \multicolumn{1}{c}{ACC ($\%$)} & \multicolumn{1}{c}{BWT ($\%$)} \\
     \midrule
    
      $0$ & $91.67\pm0.31$ & $-1.36\pm0.07$\\
      $20$ & $92.70\pm0.07$ & $-0.43\pm0.01$\\
      $40$ & $93.08\pm0.08$ & $-0.33\pm0.09$\\
      $60$ & $93.22\pm0.16$ & $-0.28\pm0.00$\\
      $80$ & $\bm{93.32\pm0.13}$ & $-0.25\pm0.00$\\
     \bottomrule
    \end{tabular}%
    }
    \end{table}

\begin{table}[!t]
    \caption{Ablation studies on the threshold ($\epsilon$) across the four benchmarks. Results are reported as mean $\pm$ standard deviation over five trials. Other hyperparameters are constant as reported in Section~\ref{sec:experimental}.}
    \label{table:ablation_reb}
    \centering
    \resizebox{0.9\columnwidth}{!}{%
    \begin{tabular}{l|c|ll}
    \toprule
     & $\epsilon$  & \multicolumn{1}{c}{ACC ($\%$)} & \multicolumn{1}{c}{BWT ($\%$)}  \\
     \midrule
    \multirow{3}{*}{S-CIFAR100} & $0.2$ & $\mathbf{77.51\pm0.18}$ & $-0.84\pm0.24$   \\
     & $0.5$ & $77.21\pm0.32$ & $-1.10\pm0.28$ \\
     & $0.8$ & $75.71\pm0.40$ & $-0.93\pm0.36$ \\
    \midrule
    \multirow{3}{*}{S-MiniImageNet} & $0.2$ & $68.61\pm0.94$ & $-1.30\pm0.18$ \\
     & $0.5$ & $\mathbf{68.83\pm0.41}$ & $-1.10\pm0.30$ \\
     & $0.8$ & $66.57\pm0.24$ & $-0.56\pm0.18$ \\
     \midrule
    \multirow{3}{*}{5-Datasets} & $0.2$ & $\mathbf{93.42\pm0.11}$ & $-0.20\pm0.06$ \\
     & $0.5$ & $93.32\pm0.13$ & $-0.25\pm0.02$ \\
     & $0.8$ & $92.28\pm0.24$ & $-0.71\pm0.18$ \\
    \bottomrule
    \end{tabular}%
    }
    \end{table}

\begin{table*}[th]
    \caption{5-Datasets statistics.}
    \label{table:datasets_stats_2}
    \centering
    \resizebox{0.8\textwidth}{!}{
    \begin{tabular}{l|ccccc}
    \toprule
    Dataset           & CIFAR10 & MNIST & SVHN & Fashion MNIST & notMNIST  \\ 
    \midrule
    Number of classes  & 10 &  10 & 10 & 10 & 10 \\
    Training samples   & 47500 &  57000 & 69595 & 57000 & 16011 \\
    Validation samples & 2500 &  3000 & 3662 & 3000 & 842 \\
    Test samples & 10000 &  10000 & 26032 & 10000 & 1873 \\
    \bottomrule
    \end{tabular}
    }
\end{table*}

\begin{table*}[th]
    \caption{List of hyperparameters used in our experiments.}
    \label{table:hyperparams}
    \centering
    \resizebox{0.8\textwidth}{!}{
    \begin{tabular}{l|ccc}
    \toprule
    Dataset &  Split CIFAR100 & Split miniImageNet & 5-Datasets \\ 
    \midrule
    Learning rate ($\eta$) &  $0.01$ & $0.1$ & $0.1$ \\
    Batch size ($b$) &  $64$ & $64$ & $64$ \\
    Batch size for conceptor comp. ($b_s$) &  $125$ & $125$ & $125$ \\
    Min. learning rate ($\eta_{th}$) &  $10^{-5}$ & $10^{-5}$ & $10^{-3}$ \\
    Learning rate decay factor       &  $1/2$ & $1/2$ & $1/3$ \\
    Patience                         &  $6$  & $6$ & $5$ \\
    Number of epochs ($E$)           &  $200$ & $100$ & $100$ \\
    Aperture ($\alpha$)              &  $6$ & $8$ & $4$ \\
    Threshold ($\epsilon$)           &  $0.5$ & $0.5$ & $0.5$ \\
    
    \bottomrule
    \end{tabular}
    }
\end{table*}
\begin{table}[th]
    \caption{Split CIFAR100 and Split miniImageNet datasets statistics.}
    \label{table:datasets_stats}
    \centering
    \resizebox{\columnwidth}{!}{
    \begin{tabular}{l|cc}
    \toprule
    Dataset           &  Split CIFAR100 & Split miniImageNet  \\ 
    \midrule
    Number of tasks ($T$) &   10 & 20               \\ 
    Sample dimensions         &      $3\times32\times32$ &   $3\times84\times84$ \\
    Number of classes per task &   10 & 5 \\
    Training samples per task &   4750 & 2375 \\
    Validation samples per task &   250 & 125 \\
    Test samples per task &   1000 & 500 \\
    \bottomrule
    \end{tabular}
    }
\end{table}

    \subsection{Model Architecture}\label{appendix:architecture}

    In this work, we utilize two models: an AlexNet-like architecture, as described in \cite{Serra2018OvercomingTask}, and a Reduced ResNet18 \cite{Lopez-Paz2017GradientLearning}.
    
    The AlexNet-like model incorporates batch normalization (BN) in every layer except the classifier layer. 
    The BN layers are trained during the first task and remain frozen for subsequent tasks. 
    The model consists of three convolutional layers with $64$, $128$, and $256$ filters, using kernel sizes of $4\times4$, $3\times3$, and $2\times2$, respectively. 
    These are followed by two fully connected layers, each containing $2048$ neurons. 
    ReLU activation functions are used throughout, along with $2\times2$ max-pooling layers after each convolutional layer. 
    Dropout is applied with rates of $0.2$ for the first two layers and $0.5$ for the remaining layers.
    
    The Reduced ResNet18 follows the architecture detailed in \cite{Saha2021GradientLearning}. 
    For the Split miniImageNet experiments, the first layer uses a stride of $2$, while for the 5-Datasets benchmark, it uses a stride of $1$.
    
    For all models and experiments, cross-entropy loss is employed as the loss function.

    \subsection{Dataset Statistics}\label{appendix:data_statistics}
    
    The statistics for the four benchmarks used in this work for continual image classification are summarized in Table~\ref{table:datasets_stats} and Table~\ref{table:datasets_stats_2}. 
    For all benchmarks, we follow the same data partitions as those used in \cite{Saha2021GradientLearning, Lin2022TRGP:Learning, Saha2023ContinualProjection}. 
    
    For the 5-Datasets benchmark, grayscale images are replicated across all RGB channels to ensure compatibility with the architecture. 
    Additionally, all images are resized to $32\times32$ pixels, resulting in an input size of $3\times32\times32$ for this benchmark.

    \subsection{Hyperparameters}\label{appendix:hyperparameters}
    The hyperparameters used in our experiments are detailed in Table~\ref{table:hyperparams}.

    \subsection{Compute resources}\label{appendix:compute_resources}
    All experiments were conducted on a shared internal Linux server equipped with an AMD EPYC 7502 32-Core Processor, 504 GB of RAM, and four NVIDIA A40 GPUs, each with 48 GB of GDDR6 memory.
    Additionally, code was implemented using Python 3.9 and PyTorch 2.2.1 with CUDA 11.8.

\end{document}